\documentclass[sigconf]{acmart}

\AtBeginDocument{%
  \providecommand\BibTeX{{%
    \normalfont B\kern-0.5em{\scshape i\kern-0.25em b}\kern-0.8em\TeX}}}

\setcopyright{acmcopyright}
\copyrightyear{2023}
\acmYear{2023}
\acmDOI{XXXXXXX.XXXXXXX}

\acmConference[EM-GIS 2023]{The 8th ACM SIGSPATIAL International Workshop on Security Response using GIS 2023}{November 13,
  2023}{Hamburg, Germany}
\acmPrice{15.00}
\acmISBN{978-1-4503-XXXX-X/18/06}

\begin{document}

\title{Causality-informed Rapid Post-hurricane Building Damage Detection in Large Scale from InSAR Imagery}


\author{Chenguang Wang}
\authornote{Equal contribution}
\email{chenguang.wang@stonybrook.edu }
\affiliation{%
  \institution{Stony Brook University}
  \streetaddress{100 Nicolls Road}
  \city{Stony Brook}
  \state{New York}
  \country{USA}
  \postcode{11794}
  }

\author{Yepeng Liu$^*$}
\email{yepeng.liu@ufl.edu }
\affiliation{%
  \institution{University of Florida}
  \streetaddress{1949 Stadium Road}
  \city{Gainesville}
  \state{Florida}
  \country{USA}
  \postcode{32611}
  }

\author{Xiaojian Zhang}
\email{xiaojianzhang@ufl.edu }
\affiliation{%
  \institution{University of Florida}
  \streetaddress{1949 Stadium Road}
  \city{Gainesville}
  \state{Florida}
  \country{USA}
  \postcode{32611}
  }

\author{Xuechun Li}
\email{xuechun.li.1@stonybrook.edu}
\affiliation{%
  \institution{Stony Brook University}
  \streetaddress{100 Nicolls Road}
  \city{Stony Brook}
  \state{New York}
  \country{USA}
  \postcode{11794}
  }

\author{Vladimir Paramygin}
\email{vladimir.paramygin@essie.ufl.edu }
\affiliation{%
  \institution{University of Florida}
  \streetaddress{1949 Stadium Road}
  \city{Gainesville}
  \state{Florida}
  \country{USA}
  \postcode{32611}
  }

\author{Arthriya Subgranon}
\email{arthriya.subgranon@essie.ufl.edu }
\affiliation{%
  \institution{University of Florida}
  \streetaddress{1949 Stadium Road}
  \city{Gainesville}
  \state{Florida}
  \country{USA}
  \postcode{32611}
  }

\author{Peter Sheng}
\email{pete@coastal.ufl.edu}
\affiliation{%
  \institution{University of Florida}
  \streetaddress{1949 Stadium Road}
  \city{Gainesville}
  \state{Florida}
  \country{USA}
  \postcode{32611}
  }

\author{Xilei Zhao}
\authornote{Corresponding author:xilei.zhao@essie.ufl.edu}
\email{xilei.zhao@essie.ufl.edu}
\affiliation{%
  \institution{University of Florida}
  \streetaddress{1949 Stadium Road}
  \city{Gainesville}
  \state{Florida}
  \country{USA}
  \postcode{32611}
  }
  
\author{Susu Xu}
\authornote{Corresponding author:susu.xu@stonybrook.edu}
\email{susu.xu@stonybrook.edu}
\affiliation{%
  \institution{Stony Brook University}
  \streetaddress{100 Nicolls Road}
  \city{Stony Brook}
  \state{New York}
  \country{USA}
  \postcode{11794}
}

\renewcommand{\shortauthors}{Wang and Liu, et al.}

\begin{abstract}
  Timely and accurate assessment of hurricane-induced building damage is crucial for effective post-hurricane response and recovery efforts. Recently, remote sensing technologies provide large-scale optical or Interferometric Synthetic Aperture Radar (InSAR) imagery data immediately after a disastrous event, which can be readily used to conduct rapid building damage assessment. Compared to optical satellite imageries, the Synthetic Aperture Radar can penetrate cloud cover and provide more complete spatial coverage of damaged zones in various weather conditions. However, these InSAR imageries often contain highly noisy and mixed signals induced by co-occurring or co-located building damage, flood, flood/wind-induced vegetation changes, as well as anthropogenic activities, making it challenging to extract accurate building damage information. In this paper, we introduced a \textit{causality-informed Bayesian network inference} approach for rapid post-hurricane building damage detection from InSAR imagery. This approach encoded complex causal dependencies among wind, flood, building damage, and InSAR imagery using a holistic causal Bayesian network. Based on the causal Bayesian network, we further jointly inferred the large-scale unobserved building damage by fusing the information from InSAR imagery with prior physical models of flood and wind, without the need for ground truth labels.  Furthermore, we validated our estimation results in a real-world devastating hurricane---the 2022 Hurricane Ian. We gathered and annotated building damage ground truth data in Lee County, Florida, and compared the introduced method's estimation results with the ground truth and benchmarked it against state-of-the-art models to assess the effectiveness of our proposed method. Results show that our method achieves rapid and accurate detection of building damage, with significantly reduced processing time compared to traditional manual inspection methods.
\end{abstract}

\ccsdesc[500]{Computing methodologies~Probabilistic reasoning}
\ccsdesc[500]{Applied computing~Physical sciences and engineering}
\ccsdesc[100]{Computing methodologies~Computer vision}

\keywords{Bayesian Networks, Geospatial Analysis, Emergency Management, Comparative Analysis}


\maketitle

\section{Introduction}
Climate change is escalating the frequency and intensity of extreme weather events, notably hurricanes, which cause massive infrastructure damage and result in staggering economic losses. For instance, the 2022 Hurricane Ian led to the evacuation of over 2.5 million people and inflicted an estimated \$113 billion in damages in the United States alone \cite{smith2021us}. In the aftermath of such disasters, rapid and accurate information on regional-scale building damage is indispensable for efficient emergency response, aiding in lifesaving activities and reconstruction budget allocations.

Traditional methods for assessing building damage in post- hurricane settings predominantly employ sensor-based inspections like street-view imagery \cite{zhai2020damage, berezina2022hurricane,ianstreetview} and aerial or satellite-based optical imagery \cite{lin2021building, calton2022using}. While ground-level inspections provide detailed damage evaluations, they are labor-intensive, time-consuming, and not easily scalable \cite{masoomi2019combined}. In contrast, optical imagery offers a quick overview but suffers limitations, particularly due to obstructive weather conditions and environmental elements such as vegetation. Additionally, optical imagery often struggles with negative transfer issues, making it less effective in different contexts without laborious manual fine-tuning \cite{wang2022avoiding}.
Emerging technologies like Interferometric Synthetic Aperture Radar (InSAR) offer promise by performing well even in adverse weather conditions \cite{barras2007satellite, lee2005application, zhao2018remote}. However, InSAR data is complex, as it amalgamates signals from various sources including geological changes, structural impairments, and human activities. Previous efforts using InSAR generally relied on historical models that were less capable of accurately discerning interconnected patterns and causal factors \cite{xu2022seismic}.

To mitigate these challenges, we introduce a ground-breaking framework that amalgamates noisy InSAR imagery with physics-driven models of hurricane impact. The framework is rooted in Bayesian Network principles and allows for a complex causal inference of interrelated factors like building damage, flooding, and wind damage. Through the use of stochastic variational inference, we transform the complex Bayesian network into an optimization problem, allowing for more accurate and timely evaluations. We substantiated the efficacy of our method through a case study of Hurricane Ian, contrasting its performance with existing methodologies.

\section{Framework}\label{sec:framework}
In this section, we present our Bayesian Network framework for improved post-hurricane multi-hazard assessment. We'll demonstrate how this probabilistic model heightens prediction accuracy by accounting for interlinked risk factors. We'll outline the workflow, methodology, and efficiency-boosting strategies, aiming to transform post-hurricane disaster management.
\begin{figure}[!h]
    \centering
    \includegraphics[width=1\linewidth]{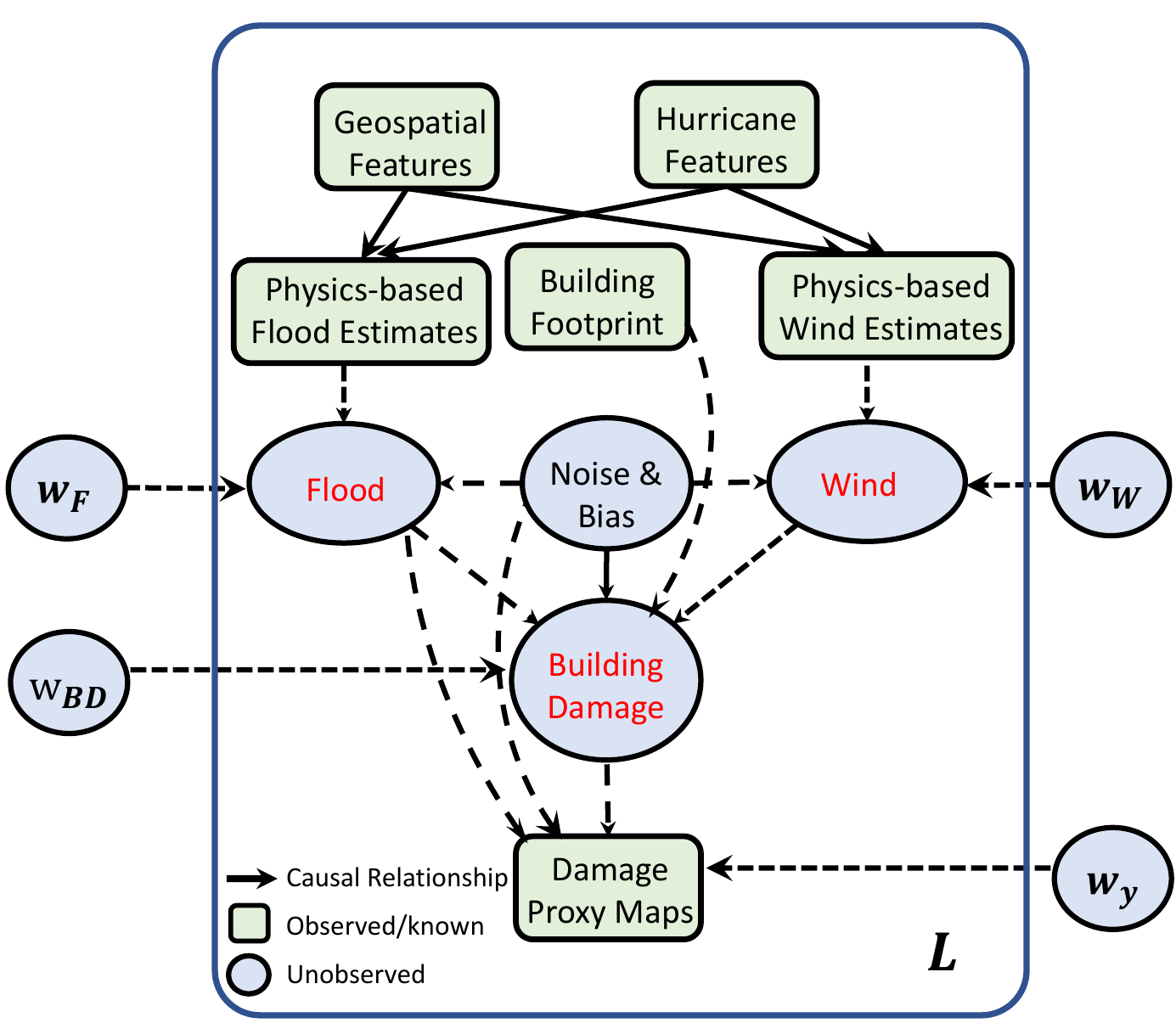}
    \caption{Overview of hurricane damage chain}
    \label{fig:overview_model}
\end{figure}
\subsection{Causal Bayesian Network Formulation for Hurricane Damage Chain }

To rigorously evaluate how hurricanes impact building integrity via secondary hazards like flooding and wind, we've devised a causal Bayesian network, shown in Figure~\ref{fig:overview_model}. This network integrates known variables, such as physics-based hazard models and building plans, with unobserved factors like the extent of flooding and building damage. The objective is to infer these latent variables using the network's causal links. Crucially, each edge in our model represents a causal, not merely correlational, relationship. We begin by mathematically formalizing these elements, converting the conceptual framework into a structured probabilistic model.

\subsubsection{\textbf{Node variables modeling}}

We designate our unobserved variables, or feature vertices, as Flood and Wind, symbolized by \(x_{F}\) and \(x_{W}\), respectively. According to prior research \cite{paleo2020vulnerability, masoomi2019combined}, these variables are continuous and follow a lognormal distribution. Building Damage is denoted as \(x_{BD}\) and treated as a binary variable, where 1 indicates severe damage or complete destruction, and 0 implies minor or no damage.Each unobserved variable \(x_{i}\) belongs to the set \(X\), where \(i \in \{W,F,BD\}\). The parent nodes for each \(x_{i}\) are contained in the set \(\mathcal{P}(i)\), pre-defined based on our understanding of the causal chain. We incorporate prior estimates \(a_{W}\) and \(a_{F}\) for wind and flooding, derived from geospatial models built on geophysical domain knowledge. Parent nodes for any sensed observations \(X\) are denoted as \(\mathcal{P}(x)\). A constant "leak node," \(x_{0i}=1\), ensures child node activity even when other parents are inactive. The Disaster Proxy Map (DPM) encapsulates the impact of multi-hazard events. Interferometric Synthetic Aperture Radar (InSAR) captures elevation changes, indicated by \(y>0\) in the DPMs. We also assume the existence of environmental noises \(\epsilon_{y}\) and \(\epsilon_{i}, i \in \{W,F,BD\}\), following a standard normal distribution.

\subsubsection{\textbf{Causal dependency modeling}}
After defining the variables, it's essential to parameterize their dependencies, or the edges connecting nodes. We use \(w_{ji}\) (\(j \in \mathcal{P}(i)\)) to represent the weight of a parent node's influence on its child node. For instance, the set of weights \(w_F\) for the Flood node incorporates the weights defining causality from its parent nodes, such as \(w_{aF}\), \(w_{\epsilon_F}\), and \(w_{0F}\).

For observed variable \(y\), we assume its mapping function from parent nodes \(\mathcal{P}(y)\) is lognormal (LN), represented as:
\begin{equation}
\log y = \sum_{k\in\mathcal{P}(y)}w_{ky}x_{k} + w_{\epsilon_{y}}\epsilon_{y} + w_{0y} 
\end{equation}
Here, \(\mathcal{P}(y) = \{x_{F}, x_{BD}, \epsilon_y, x_{0}\}\).

For unobserved root variables \(i \in \{W, F\}\), their parent nodes include prior estimates \(a_i\), environmental noises \(\epsilon_i\), and the leak node \(x_0\). Their mapping function is:
\begin{equation}
\log x_i = \sum_{k\in\mathcal{P}(i)}w_{ki}x_{k} + w_{\epsilon_{i}}\epsilon_{i} + w_{0i} 
\end{equation}
This aligns with our assumption that \(x_{F}\) and \(x_{W}\) follow a lognormal distribution.

For Building Damage (\(i=BD\)), \(\mathcal{P}(BD) = \{x_{F}, x_{W}\}\), and its activation function is:
\begin{equation}
    \log\frac{p(x_{i} = 1|\mathcal{P}(i),\epsilon_{i})}{1-p(x_{i} = 1|\mathcal{P}(i),\epsilon_{i})} = \sum_{k\in \mathcal{P}(i)}w_{k,i}x_{k} + w_{\epsilon_{i}}\epsilon_{i} + w_{0i}.
\end{equation}
Given that the causal coefficients (represented by dashed arrows in Figure~\ref{fig:overview_model}) and unobserved variables (blue circles) are unknown, the challenge lies in inferring this complex multi-layer Bayesian network, which combines unknown continuous and discrete variables along with uncertain dependency parameters.

\subsection{Stochastic Variational Inference}
After constructing the Bayesian network, we face two main challenges: the unknown parameters for causal dependencies and hazard distributions, and computational constraints for high-resolution mapping of large areas. To overcome this, we implement variational inference. This computational technique allows us to factorize the Bayesian network and approximate the posterior distributions of unobserved variables, such as building damage and secondary hazards, by maximizing the log-likelihood of the observed data. To ensure the method's scalability, we execute the inference on a small, randomly selected batch of locations during each iteration. This balances computational efficiency with the need for comprehensive spatial analysis. The core objective of the variational inference is to derive and maximize a tight lower bound on the log-likelihood, which is formulated as a function of both the posteriors of unobserved variables and the causal weights. This approach effectively translates the theoretical Bayesian network into a practical, computationally efficient mechanism for large-scale hazard assessment.
Accordingly, we derive the variational lower bound is :
\begin{equation}
\begin{split}
    \log p(Y)    & \geq  \sum_{l\in N}\{\underbrace{\mathbb{E}_{q(X^{l},\epsilon^{l})}[\log p(y^{l},X^{l},\epsilon^{l})]}_{[1]} - \underbrace{\mathbb{E}_{q(X^{l},\epsilon^{l})}[\log q(X^{l},\epsilon^{l})]}_{[2]}\}
\end{split}
\end{equation}
\noindent where $q$ is the posteriors of all unknown variables. 

To further obtain the explicit form of the final variational bound, we expand the \textbf{item[1]} as:
\begin{equation}
\begin{split}
 & \mathbb{E}_{q(X^{l},\epsilon^{l})}[\log p(y^{l},X^{l},\epsilon^{l})]\\
& = \underbrace{\mathbb{E}[\log p(y|\mathcal{P}(y),\epsilon_{y})]}_{[3]} + \underbrace{\sum_{i}\mathbb{E}[\log p(x_{i}|\epsilon_{i},\mathcal{P}(i))]}_{[4]} \\
&+ \underbrace{\sum_{i}\mathbb{E}[\log p(\epsilon_{i})] + \mathbb{E}[\log p(\epsilon_{y})] }_{C_{1}}
\label{item1}
\end{split}
\end{equation}
Since $y|\mathcal{P}(y) \sim LN(\sum_{k\in\mathcal{P}(y)}w_{ky}x_{k}  + w_{0y}, w_{\epsilon_{y}}^{2})$, where $\mathcal{P}(y) = \{F,BD\}$. So we can calculate \textbf{item[3]}: 
\begin{equation}
\begin{split}
&\mathbb{E}[\log p(y|\mathcal{P}(y),\epsilon_{y})] \\
& = -\log y - \log |w_{\epsilon_{y}}| - \frac{(\log y)^{2} + w_{0y}^{2} - 2w_{0y}\log y}{2w_{\epsilon_{y}}^{2}} \\
& + \frac{\sum_{k \in \mathcal{P}(y)}w_{ky}^{2}\mathbb{E}(x_{k}^{2}) + 2(w_{0y} - \log y)\sum_{k\in\mathcal{P}(y)}w_{ky}\mathbb{E}(x_{k})}{2w_{\epsilon_{y}}^{2}}
\end{split}
\end{equation}

\noindent where $k=F$: $E(x_k) = \exp(\mu_k+\sigma^2/2)$ and $E(x_k^2) = (\exp(\sigma^2)-1)\exp(2\mu_k+\sigma^2)+\exp(\sigma^2+2\mu_k)$. ; For $k=BD\}$, $\mathbb{E}(x_{i}) = q_k, \mathbb{E}(x_{i}^{2}) = q_k$.

For the \textbf{item[4]}, It should be derived separately because the $i \in \{W,F\}$ and $i \in \{SD,RD\} $ are different situations here, so the equation could be expanded first into:$\sum_{i}\mathbb{E}[\log p(x_{i}|\epsilon_{i},\mathcal{P}(i))] = \sum_{i\in \{W,F\}}\mathbb{E}[\log p(x_{i}|\epsilon_{i},\mathcal{P}(i))] + \sum_{i\in \{RD,BD\}}\mathbb{E}[\log p(x_{i}|\epsilon_{i},\mathcal{P}(i))]$
Then based on this equation, we could divide the equation into two different scenarios.

When the unobserved variable is continuous, we will derive the $ \mathbb{E}[\log p(x_{i}|\epsilon_{i},\mathcal{P}(i))] $ for $i \in \{W,F\}$. Since $x_{i}|\epsilon_i$ follows a lognormal distribution, then $\log x_{i}|\epsilon_i$ follows a normal distribution. Then we could get the \textbf{item[4]} for $i \in \{W,F\}$:
\begin{equation}
\mathbb{E}[\log p(x_{i}|\epsilon_{i},\mathcal{P}(i))] = -(w_{a_{i}}a_{i} + w_{0i}) - \log|w_{\epsilon_{i}}| 
\end{equation}
When an unobserved variable $i$ is discrete, for example, a binary distribution, we can determine the result of $ \mathbb{E}(\log p(x_{i} = m_{i}|\mathcal{P}(i),\epsilon_{i})) $ for $i \in \{BD\}$. The computation of \textbf{item[4]} for $i \in \{BD\}$ is :
\begin{equation}
\begin{split}
&\mathbb{E}(\log p(x_{i}|\mathcal{P}(i),\epsilon_{i})) \\
&=q_i\mathbb{E}(-\log[1+\exp(-w_{\epsilon_i}\epsilon_i-w_{0i}-\sum_{k\in \mathbf{P}(i)}w_{ki}x_k)])\\
    &+(1-q_i)\mathbb{E}(-\log [1+\exp(w_{\epsilon_i}\epsilon_i+w_{0i}+\sum_{k\in \mathbf{P}(i)}w_{ki}x_k)]).
\end{split}
\end{equation}
However, the distribution of $-\log [1+\exp(w_{\epsilon_i}\epsilon_i+w_{0i}+\sum_{k\in \mathbf{P}(i)}w_{ki}x_k)$ is intractable as it is a log-sum-exp of mixing a series of discrete variables and continuous variables. Therefore, we need to get a tight lower bound of its expectation. Here without the loss of generality, we start from the case where $i$ has a single active parent. 

With multivariate Taylor expansion, we can apply the standard quadratic bound for $\text{log}(1+\text{exp})$ \cite{jaakkola1997variational}:
\begin{equation*}
       \text{log}(1 + e^{z})  \leq g(\gamma)(z^{2} - \gamma^{2}) + \frac{z - \gamma}{2} + \text{log}(1 + e^{\gamma}) 
\end{equation*}
where $\gamma\in (0, \infty), g(\gamma) = \frac{1}{2\gamma}[\frac{1}{1+e^{-\gamma}} - \frac{1}{2}]$. Thus, we can obtain the lower bound as:

\begin{equation}
\begin{split}
&\mathbb{E}(\log p(x_{i}|\mathcal{P}(i),\epsilon_{i})) \notag \\
&\geq (1-q_i)\mathbb{E}(z_{i}) - \{ g(\gamma_{i})(\mathbb{E}(z_{i}^{2})  - \gamma_{i}^{2}) + \frac{\mathbb{E}(z)  - \gamma_{i}}{2} + \log(1+e^{\gamma_{i}})\}
\end{split}
\end{equation}

\noindent where
\begin{equation}
\begin{split}
&\mathbb{E}(z) = \sum_{k\in \mathcal{P}(i)}w_{ki}\mathbb{E}(x_{k}) + w_{0i}\\
&\mathbb{E}(z^{2}) = \sum_{k\in \mathcal{P}(i)}w_{ki}^{2}\mathbb{E}(x_{k}^{2}) + w_{\epsilon_{i}}^{2} + w_{0i}^{2} + 2w_{0i}\sum_{k\in \mathcal{P}(i)}w_{ki}\mathbb{E}(x_{k}) \\
&+ \sum\limits_{\substack{r,s \in \mathcal{P}(i) \\ r\neq s}}w_{ri}w_{si}\mathbb{E}(x_{r})\mathbb{E}(x_{s})
\end{split}
\end{equation}
It is noted that for $k\in\{W,F\}$, as they are leaf nodes with all parent nodes known, and we postulate that $x_{k}$ follows a lognormal posterior with parameters of $\mu_{k}$ and $\sigma_{k}$, there is $\mathbb{E}(x_k) = \exp(\mu_k+\sigma^2/2)$ and $\mathbb{E}(x_k^2) = (\exp(\sigma^2)-1)\exp(2\mu_k+\sigma^2)+\exp(\sigma^2+2\mu_k)$.
After we finish the all the calculation form the \textbf{item[1]}, we further move to derive the expansion of the \textbf{item[2]} as: 
\begin{equation}
\begin{split}
\mathbb{E}_{q(X^{l},\epsilon^{l})}[\log q(X^{l},\epsilon^{l})] & = \sum_{i \in \{\}}q_{i}\log q_{i}+(1-q_i)\log(1-q_i) \\
& - \sum_{i \in \{F,W\}}[\mu_{i} + \log \sigma_{i}]+C_{2} .
\end{split}
\end{equation}
\noindent where $C_2$ is a fixed constant.

With the derivation of a tight variational lower bound for our causal Bayesian network focused on hurricane-induced hazards and damage, we proceed to optimize both unobserved variable posteriors and causal coefficients. Utilizing an Expectation-Maximization (EM) framework, the expectation step involves formulating closed-form update equations for the posteriors of wind (W), flood (F), and building damage (BD) by setting the gradients of the lower bound to zero. During the maximization phase, we employ Stochastic Gradient Descent (SGD) with mini-batches of randomly sampled data points from diverse locations. We've customized Stochastic Variational Inference to fast-track computation across expansive, high-resolution maps. In this, edge weights for the next iteration $t + 1$ are updated as follows: 
\begin{equation}
 \textbf{w}_{(t+1)} = \textbf{w}_{(t)} + \rho A \nabla \mathcal{L}(\textbf{w})
\end{equation}
where $\rho$ adjusts the learning rate and $A$, serving as a preconditioner, is configured as the identity matrix to encourage model convergence.

\noindent\textbf{Local pruning strategy for computational efficiency}: To address computational challenges in large-scale mapping, we've created a local pruning strategy for our causal graph. Based on the sparsity of the real-world casual graph, we remove inactive nodes—like those for building damage in a location without any building footprints —while keeping crucial active nodes for parameter updates.

\section{A Case Study of Hurricane Ian}

The 2022 Hurricane Ian \cite{https://doi.org/10.17603/ds2-3pc2-7p82} is chosen as the case study. This section introduces the background of Hurricane Ian, and provides a detailed description of street-view imagery data collection, data pre-processing and damage level annotation.

\subsection{Background of Hurricane Ian}

On September 28, 2022, Hurricane Ian struck Lee County, FL, displacing over 130,000 residents and causing extensive coastal erosion\footnote{\url{https://ianprogress.leegov.com/}}. Our study involved capturing street-view images of 2,472 affected buildings from both StEER's Hurricane Ian Response\footnote{\url{https://www.steer.network/hurricane-ian}} and our own field investigations. Utilizing StEER's FAST handbook\cite{kijewski2019field}, two expert annotators classified the exterior damage into five levels, from 0 (no or minor damage) to 4 (destroyed). Statistical tests indicated high inter-coder reliability, with a Krippendorff's alpha of 0.86 and a correlation coefficient of 0.97 (p < 0.001). The final dataset revealed 6.1\% of buildings suffered severe-to-total damage, 72.2\% had no visible damage, and the remaining 27.8\% were damaged. 
\begin{figure}[!h]
    \centering
    \includegraphics[width=0.95\linewidth]{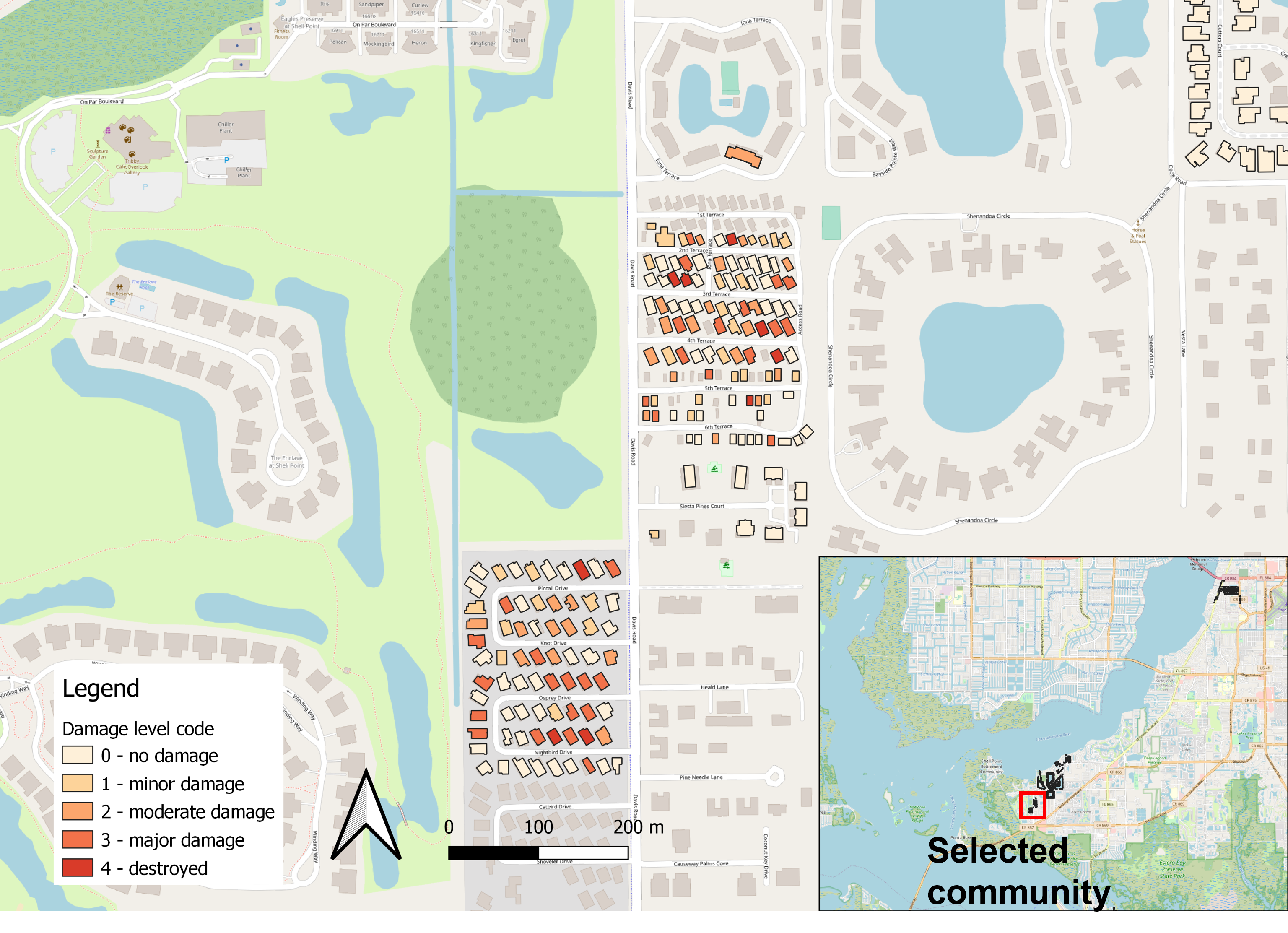}
    \caption{An example of damage level-labeled parcels in a selected community}
    \label{fig:annotation_example}
\end{figure}

\subsection{Datasets and Evaluation}
\begin{figure*}[!h]
    \includegraphics[width=\textwidth]{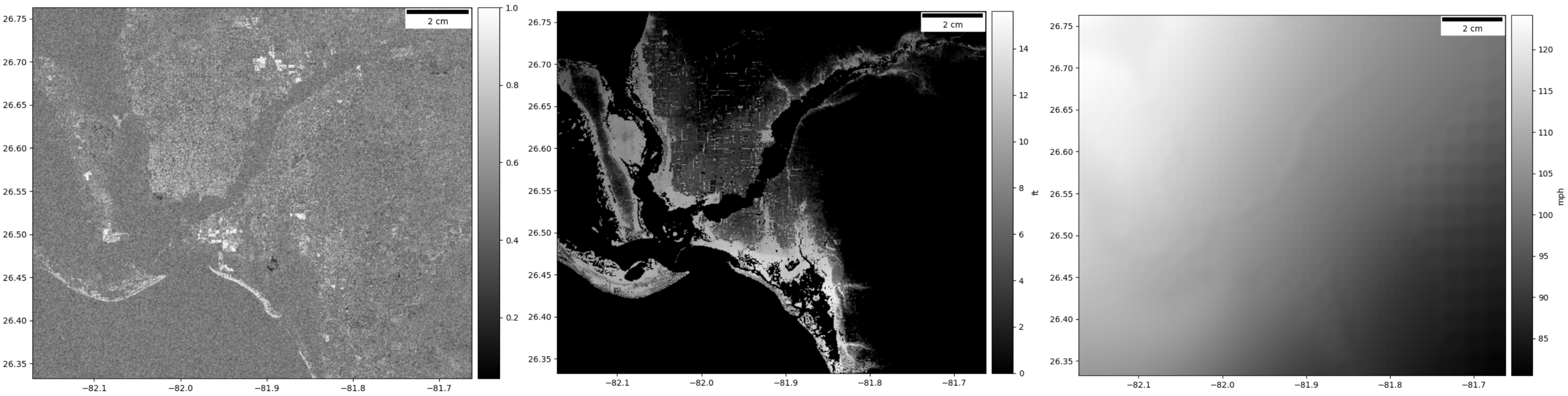}
    \vspace{-0.3cm}
    \caption{Overview of the DPM, Flood Map and Wind Map in Lee County, FL (from left to right)}
    \label{fig:overview_1}
\end{figure*}

Our framework, depicted in Figure~\ref{fig:overview_model}, heavily relies on three key inputs for inference: \textbf{Wind Map}, \textbf{Flood Map}, and \textbf{Damage Proxy Map}. Crafted by NIST and ARA\footnote{\url{https://www.nhc.noaa.gov/archive/2022/IAN_graphics.php}}, the \textbf{Wind Map} employs advanced simulations to detail peak and sustained winds. The \textbf{Flood Map}, produced via the CH3D-SWAN model\cite{sheng2010simulation}, offers storm surge simulations in Florida counties with a 20-meter grid granularity, and it is calibrated with hurricane Ian's wind data\cite{peter2022sensitivity}. Meanwhile, the \textbf{Damage Proxy Map} (DPM) is a crucial contribution from NASA's ARIA team, showcased in Fig. \ref{fig:overview_1}. Derived from Copernicus Sentinel-1 SAR images\cite{NASA_ARIA_2022}, the DPM provides reliable, cloud-penetrant data in an 85 by 76 kilometer area around Fort Myers. Leveraging InSAR technology, the DPM serves as a foundational grid for rasterizing our datasets.

In our study, we use the Receiver Operating Characteristic (ROC) curve and its Area Under the Curve (AUC) for evaluation. The ROC curve plots True Positive Rate (TPR) against False Positive Rate (FPR). A high AUC near 1 signifies strong model separability in identifying true damages versus false predictions. We set a threshold to categorize outcomes as "damage exists" or "no damage," allowing for practical performance comparison. The ROC curve offers a balanced assessment over varied thresholds from 0 to 1.

\subsection{Benchmarks}

We evaluate our framework against three types of benchmark methods: Optical Imagery-Based Models, Fragility Curve-Based Models, and Direct DPM-Based Models. These benchmarks offer comprehensive comparisons to traditional approaches in post-hurricane building damage detection.

\textbf{Optical Imagery-Based Models}: We employ two top-performing models from the xView2 Challenge\footnote{\url{https://www.xview2.org/}} as baselines: FCS-Net\cite{khvedchenya2021fully} and Dual-HRNet\footnote{\url{https://github.com/DIUx-xView/xView2_fifth_place/blob/master/figures/xView2_White_Paper_SI_Analytics.pdf}}. FCS-Net uses a Siamese architecture with ResNet34\cite{he2015deep} and U-Net \cite{ronneberger2015unet}components, while Dual-HRNet integrates two HRNets \cite{wang2020deep, sun2019highresolution} with serveral fusion blocks. We first test these pre-trained models on our datasets without fine-tuning, and then fine-tune them for further evaluation. The results are summarized in Table~\ref{table_combined}.

\textbf{Fragility Curves}: We leverage a fragility curve model developed by Andres Paleo-Torres et al.\cite{paleo2020vulnerability}, focusing on the vulnerability of Florida's residential buildings to coastal flooding. It adapts tsunami fragility functions and outlines six damage states, offering a tailored benchmark to measure hurricane-induced building vulnerability in Florida.

\subsection{Results}
In this section, we present the results derived from our model and draw comparisons with conventional methodologies, including the flood-map-based model, DPM-based model, and optical imagery-based models. This comparative analysis offers a clear understanding of our model's performance, highlighting its advantages and potential improvements over traditional techniques.

\subsubsection{Result analysis}
Our model excels in key performance metrics, making it a compelling alternative to more traditional, label-dependent methods. With a True Positive Rate (TPR) of 0.8293 and an Area Under the Curve (AUC) score of 0.7553, it surpasses other models in the landscape. While its True Negative Rate (TNR) of 0.6221 may not be the highest, it is nonetheless a competitive figure. Importantly, the model obviates the need for ground-truth labels, a feature that greatly simplifies the training process and saves significant time and resources. This is especially advantageous in scenarios where label acquisition is cumbersome or not feasible.

Compared to this, the fine-tuned Dual-HRNet model, which necessitates labels for training, produced TPR and TNR scores of 0.8217 and 0.6251, respectively—figures that are closely matched but not exceeding ours. Other benchmark models, such as the Pure DPM-based Estimation and the Fragility Curve-based Estimation, lagged behind in both TPR and TNR. Additionally, the fact that our model performs robustly without requiring additional flood or wind maps highlights its adaptability and widens its applicability. Therefore, our model not only streamlines the training process but also demonstrates that label-free, efficient training can yield results on par with more resource-intensive approaches.

\begin{table}[htbp]
\centering

\caption{Comparison of TPR, TNR, and AUC of our model and baselines. Note that (1) our probabilistic model that fuses DPMs and physics-based flood and wind model does not require any labels in training stage, (2) FCS-Net and Dual-HRNet are computer vision models using optical satellite imagery and pretrained in historical events. "w/o finetuning" means the model is directly adopted to estimate building damage, which is equivalent to our model setup, "finetuned" means the model is finetuned using labeled building damage datasets for Hurricane Ian. The results show that our model that does not require any labeled data achieves comparable performance as the model finetuned on labeled data. NA means Not Available, meaning AUC is not calculated for deterministic models like FCS-Net and Dual-HRNet.}

\label{table_combined}
\resizebox{0.49\textwidth}{!}{%
\begin{tabular}{@{}lccc@{}}
\toprule
Model & TPR & TNR & AUC \\ \midrule
\textbf{Our model (w/o label training)} & \textbf{0.8293} & \textbf{0.6221}& \textbf{0.7553}\\
FCS-Net (w/o finetuning) & 0.2713 & 0.8941 &NA\\
FCS-Net (w/ finetuning) & 0.2098 & 0.9386 &NA\\
Dual-HRNet (w/o finetuning) & 0.0912 & 0.9795 &NA\\
Dual-HRNet (w/ finetuning) & 0.8217 & 0.6251 &NA\\
DPM-based Model (w/o label training) & 0.6498 & 0.6249 &0.6739\\
Fragility Curve (w/o label training) & 0.5669 & 0.6246 &0.5695\\
\bottomrule
\end{tabular}%
}
\end{table}

\subsubsection{Ablation Study}

Our ablation study yielded key insights into the interplay between different components and parameters in our model. First, we noted that batch size has a marked influence on both computational time and model performance, as evidenced by Table \ref{table:batchtime}. Specifically, increasing the batch size from 128 to 1024 slashed computational time from 5812 to 941 seconds but resulted in a modest performance decline, as AUC dropped from 0.7553 to 0.7311. This highlights a trade-off between computational speed and predictive accuracy. Second, we assessed the contributions of Variational Inference (VI) and Pruning strategies, as outlined in Table \ref{table:NVLB}. VI Full exhibited the best performance with an AUC of 0.7825 and a Variational Lower Bound (VLB) of 1.1442, outperforming the worst performer, MCMC Local, which posted an AUC of 0.7117 and a VLB of 0.7140. This stark contrast substantiates the superiority of VI methods over MCMC methods. Further, within both VI and MCMC categories, the Local Pruned Graph yielded slightly lower AUC but better VLB, suggesting it offers a tighter lower bound at a minimal AUC cost. Thus, our ablation study not only quantifies the impact of batch size and computational techniques but also validates the strategic choices made in our model's architecture.

\vspace{-0.3cm}
\begin{table}[htbp]
\begin{center}
\caption{The computational time  for different batch sizes.}
\vspace{-0.3cm}
\scalebox{1}{
\begin{tabular}{c|cccc}
\toprule
\textbf{Batch size $m$}	& \textbf{128} & \textbf{256} & \textbf{512}	& \textbf{1024} \\
\hline
AUC &  0.7553  & 0.7553   &  0.7329 &  0.7311\\
Time used (second) & 5812& 2729  & 1267  &941 \\
 \toprule
\end{tabular}
}
\label{table:batchtime}
\end{center}
\end{table}
\vspace{-0.2cm}
\begin{table}[htbp]
\begin{center}
\caption{Evaluation of the effectiveness of Variational Inference and Prunning strategy in our framework. We present AUC values and variational lower bound (VLB).}
\vspace{-0.3cm}
\scalebox{1}{
\begin{tabular}{c|cccc}
\toprule
\textbf{Method}  & AUC & VLB\\
\hline
VI Full &0.7825& 1.1442\\
VI Local &0.7553 & 1.1249 \\
MCMC Full & 0.7271 & 0.7797 \\
MCMC Local & 0.7117 & 0.7140 \\
 \toprule
\end{tabular}
}
\label{table:NVLB}
\end{center}
\end{table}
\vspace{-0.3cm}


\section{Discussion and Conclusion}

This paper unveils a novel deep learning model that uses a causal Bayesian network to quickly assess building damage post-hurricane using InSAR imagery. Tested on Hurricane Ian, our model outperforms both state-of-the-art optical imagery benchmarks and traditional methods, all without requiring ground-truth labels for rapid deployment post-disaster.

The strength of our model lies in its causal graph, which simulates real-world hurricane-induced hazards leading to building damage, unlike previous models that merely map noisy remote sensing data to damage outcomes. This allows for the effective fusion of physics-based flood and wind estimates with sensor data, enhancing accuracy and speed of assessments. However, existing optical imagery models like FCS-Net and Dual-HRNet falter when applied to new events, demonstrating poor generalizability and low True Positive Rates (Table \ref{table_combined}). Fine-tuning these models presents logistical hurdles, primarily the need for labor-intensive ground-truth labeling, limiting their usefulness in rapid damage assessments.

While promising, our model does have limitations, particularly in areas with steep slopes that can interfere with SAR backscatter. Future work should consider integrating InSAR with optical imagery for a more comprehensive and accurate damage assessment.

\begin{acks}
C. W., X. L. and S. X. are supported by U.S. Geological Survey Grants (G22AP00032,G23AS00249) and Stony Brook University. 
X. Zhao and X. Zhang are supported by the National Science Foundation under Grant No. 2303578 and by an Early-Career Research Fellowship from the Gulf Research Program of the National Academies of Sciences, Engineering, and Medicine. 
P. Sheng and V. Paramygin are supported by a cooperative agreement funded by the National Oceanic and Atmospheric Administration's ESLR Program under award NA19NOS4780178 to the University of Florida.
Any opinions, findings, and conclusions or recommendations expressed in this material are those of the authors and do not necessarily reflect the views of the funders.
\end{acks}

\bibliographystyle{ACM-Reference-Format}
\bibliography{reference}

\appendix

\end{document}